 \documentclass[smallabstract,smallcaptions]{dccpaper}

\usepackage{epsfig}
\usepackage{citesort}
\usepackage{amsmath}
\usepackage{amssymb}
\usepackage{color}
\usepackage{url}

\newlength{\figurewidth}
\newlength{\smallfigurewidth}

\begin{document}

\title
{\large
\textbf{TLIC: Learned Image Compression with ROI-Weighted Distortion and Bit Allocation}
}

\author{%
Wei Jiang$^{1}$, Yongqi Zhai$^{1,2}$, Hangyu Li$^1$, Ronggang Wang$^{1,2,3*}$\thanks{* Corresponding author.}\\[0.5em]
{\small\begin{minipage}{\linewidth}\begin{center}
\begin{tabular}{ccc}
$^{1}$Shenzhen Graduate School, Peking University\\
$^{2}$Pengcheng Laboratory\quad \quad  $^{3}$Migu Culture Technology Co., Ltd\\
\url{jiangwei@stu.pku.edu.cn}\quad \quad \url{rgwang@pkusz.edu.cn}
\end{tabular}
\end{center}\end{minipage}}
}

\maketitle
\thispagestyle{empty}

\begin{abstract}
This short paper describes our method for the track of image compression. To achieve better perceptual quality, we use the adversarial loss to generate realistic textures, use region of interest (ROI) mask to guide the bit allocation for different regions. Our Team name is TLIC.
\end{abstract}
\section{Introduction}
Learned image compression~\cite{balle2018variational} becomes an active area in recent years.
Some of the models~\cite{cheng2020learned,he2022elic,jiang2022mlic,jiang2023mlic,jiang2023slic} have outperformed latest non-learned codec VVC. Most of the methods focus on 
non-perceptual metric optimization. To improve perceptual quality, 
some methods~\cite{agustsson2019generative,mentzer2020high,agustsson2023multi} employ Generative Adversial Networks (GANs)~\cite{goodfellow2014generative} to generate perceptual
textures. In addition, VGG\cite{simonyan2014very} or LPIPS\cite{zhang2018unreasonable}-based learned metrics are employed to help convergence.
Since people have different standards for subjective criterion, for some content sensitive images, such as faces, documents, keeping its authenticity is more important than generating vivid but fake texture. To this end, in this paper,
based on the generative image compression framework~\cite{mentzer2020high},
we propose Learned Image Compression with ROI-Weighted Distortion and Bit Allocation, named TLIC. We employ ROI to adjust the weight of the symbols in the latent to allocate more bits to regions of interest.
In addition, to enhance the guidance, we employ roi mask to control the weight of distortion of each pixels.
At last, to satisfy target bits, our model is trained with a variable rate compression method inspired from \cite{cui2021asymmetric}.
\section{Method}
\subsection{Overview}
In this section, we briefly introduce our method for perceptual image compression. 
The architecture of our TLIC is similar to Ma~\textit{et al.}~\cite{ma2021variable}.
The gain and inverse gain units~\cite{cui2021asymmetric} are employed for variable rate control.\par
The optimization process of our TLIC contains two stages. In the first stage,
the model is optimized for mean-square-error (MSE) with gain and inverse gain units.
During the sencond stage, the model is optimized for perceptual quality.\par
For perceptual quality optimization, we adopt adversarial loss~\cite{goodfellow2014generative}
\cite{agustsson2019generative}\cite{mentzer2020high} to guide the decoder to generate realistic textures at low bit-rate. Our discriminator employs the Unet~\cite{ronneberger2015u} architecture. In addition to adversarial loss, we used L1 loss to keep the texture sharp. LPIPS loss~\cite{zhang2018unreasonable} and Style loss~\cite{gatys2016image} are also used to enhance the quality. We also use the Laplacian loss~\cite{niklaus2018context} to reduced color variation. Following \cite{ma2021variable}, we use a region of interest (ROI) mask to guide the network to allocate more bits in regions of interest. The overall loss function is 
\begin{equation}
\begin{split}
    \mathcal{L} = \lambda_r \times \mathcal{R} + \lambda_{mse} \times (\delta \odot \mathcal{D}_{mse}) + \lambda_{L1}^{ROI} \times (\delta \odot \mathcal{D}_{L1}) +\\ \lambda_{L1}^{non-ROI} \times [(1 - \delta) \odot \mathcal{D}_{L1}] + \lambda_{lpips} \times \mathcal{D}_{lpips} + \\\lambda_{sty} \times \mathcal{D}_{sty} + \lambda_{lap} \times \mathcal{D}_{lap} + \lambda_{adv} \times \mathcal{D}_{adv},
\end{split}
\end{equation}
where $\mathcal{R}$ is the rate, $\mathcal{D}_{mse}$ is the MSE distortion, $\mathcal{D}_{L1}$ is the L1 distortion, $\mathcal{D}_{lpips}$ is the VGG-16~\cite{simonyan2014very} Lpips~\cite{zhang2018unreasonable} distortion, $\mathcal{D}_{sty}$ is the style loss~\cite{gatys2016image}, $\mathcal{D}_{lap}$ is the Laplician distortion~\cite{niklaus2018context}, $\mathcal{D}_{adv}$ is the BCE adversarial distortion, $\delta$ is the ROI map. We use $\{\lambda_r, \lambda_{mse}, \lambda_{L1}^{ROI}, \lambda_{L1}^{non-ROI}, \lambda_{lpips}, \lambda_{sty}, \lambda_{lap}, \lambda_{adv}\}$ to adjust the weight of each loss.

\subsection{ROI-Weighted Distortion and Bit Allocation}
In our method, following Ma~\textit{et al.}~\cite{ma2021variable}, we employ saliency map as the ROI map because the 
saliency detection can distinguish the image into the focused area and background, which is more suitable to our strategy.
We employ the RMformer~\cite{deng2023recurrent} to generate ROI maps.
The RMformer is fixed during training and testing. The process is formulated as:
\begin{equation}
Mask_{2D} = \textrm{sigmoid}(\textrm{RMformer}(\boldsymbol{x})),
\label{equation1}
\end{equation}
where $\boldsymbol{x}$ is the input image, $Mask_{2D}$ is the detected ROI map.
To 
The $Mask_{2D}$ is further pooled to smooth the boundaries for smooth bit-allocation.
\begin{equation}
RM_{2D} = \textrm{AvgPool}(Mask_{2D}),
\label{equation2}
\end{equation}
where $RM_{2D}$ is the smoothed ROI map.
In addition, we use $\alpha$ to control the weight of the ROI in terms of rate allocation. What's more, we protect a certain number of channels to retain appropriate information for the background to avoid the fading of its reconstructed texture.
Finally, the bit-allocation is achieved at the encoder side and can be formulated as:
\begin{equation}
\begin{aligned}
    \boldsymbol{y} &= g_a(\boldsymbol{x}),\\
\tilde{\boldsymbol{y}} &= \boldsymbol{y} \odot \frac{RM_{2D} + \alpha}{1+\alpha},\\
\overline{\boldsymbol{y}} &= \boldsymbol{y}_{0-47}||\tilde{\boldsymbol{y}}_{ch48-191},\\
\hat{\boldsymbol{y}} &= Q(\overline{\boldsymbol{y}}),\\
\hat{\boldsymbol{x}} &= g_s(\hat{\boldsymbol{y}}),
\end{aligned}
\label{equation3}
\end{equation}
where $g_a$ is the analysis transform, $g_s$ is the synthesis transform.
To guide the model to allocate more bits on ROI regions and L1 and MSE are pixel-wise losses, we directly use the ROI map to adjust the weight of each pixel.
\subsection{Adversial Training}
To generate realistic textures, we employ the adversarial loss to optimize the model. Following existing methods~\cite{mentzer2020high,ma2021variable}, 
we employ BCE loss. The loss function is formulated as:
\begin{equation}
    \begin{aligned}
        \mathcal{D}_{adv} &= \mathbb{E}(-\log(D(\hat{\boldsymbol{y}}, \hat{\boldsymbol{x}})),\\
        \mathcal{D}_{disc} &= \mathbb{E}(-\log(1-D(\hat{\boldsymbol{y}}, \hat{\boldsymbol{x}})) + \mathbb{E}(-\log(D(\hat{\boldsymbol{y}}, \boldsymbol{x})),
    \end{aligned}
\end{equation}
where $D$ is the discriminator. $\mathcal{D}_{disc}$ is employed to optimize the descriminator.
We employ the U-net discriminator~\cite{ronneberger2015u} for more accurate pixel-wise feedback while maintaining syntax feedback.
\subsection{Variable Rate Adaptation}
The gain units and inverse gain units are employed for continuous rate adaptation.
Specifically, the model is trained for $n$ target bit-rates. The gain units $M_{G}\in \mathrm{R}^{c\times n}$ is employed to adjust the quantization step of each channel,where $c$ is the channel number of each channel. The inverse gain is $1/M_{G}$ in our model.
The process in Equation~\ref{equation3} becomes
\begin{equation}
\begin{aligned}
    \boldsymbol{y} &= g_a(\boldsymbol{x}),\\
\tilde{\boldsymbol{y}} &= \boldsymbol{y} \odot \frac{RM_{2D} + \alpha}{1+\alpha},\\
\overline{\boldsymbol{y}} &= \boldsymbol{y}_{0-47}||\tilde{\boldsymbol{y}}_{ch48-191},\\
\hat{\boldsymbol{y}} &= Q(\overline{\boldsymbol{y}} \odot M_{G}),\\
\hat{\boldsymbol{x}} &= g_s(\hat{\boldsymbol{y}} \odot \frac{1}{M_{G}}).
\end{aligned}
\label{equation5}
\end{equation}
\subsection{Entropy Model}
The entropy model is a simplified version of recent linear complexity multi-reference entropy model~\cite{jiang2023mlic}.
The latent residual prediction modules~\cite{minnen2020channel} and 
the inter-slice global context modules are removed 
to reduce model parameters and we employ one layer convolution-based local context modules. The number of slices are set to $5$. 
\subsection{Training}
To 
accelerate the training, $320\times 320$ patches are employed during training and 
the batch size is set $8$.
\section{Conclusion}
In this report, we propose to employ the ROI-weighted rate allocation and distortion for better perceptual quality. To enhance the pixel feedback and syntax feedback of adversarial optimization, we employ the U-net based discriminator architecture. To achieve the target bit-rate, we employ the gain units and inverse gain units.
\Section{References}
\bibliographystyle{IEEEbib}
\bibliography{refs}

\end{document}